\documentclass[letterpaper]{article} 
\usepackage[preprint]{aaai2027}  
\usepackage[hyphens]{url}  
\usepackage{graphicx} 
\usepackage{natbib}  
\usepackage{caption} 
\usepackage{algorithm}
\usepackage{algorithmic}
\usepackage{soul}
\usepackage{amsmath}
\usepackage{newfloat}
\usepackage{listings}
\DeclareCaptionStyle{ruled}{labelfont=normalfont,labelsep=colon,strut=off} 
\floatstyle{ruled}
\newfloat{listing}{tb}{lst}{}
\floatname{listing}{Listing}

\usepackage{booktabs}

\title{On the Depth Scalability of Logic Gate Networks}
\author{
    Taegun An, Dohun Kim, Haebeom Lee, Changhee Joo*
}
\affiliations{

    Jung Woonoh IT \& General Education Center, Korea University 145,\\
    Anam-ro, Seongbuk-gu, Seoul 02841, Republic of Korea\\
    \{antaegun20, dohunkim, haebeomlee, changhee\}@korea.ac.kr
}

\begin{document}

\maketitle

\begin{abstract}
Logic Gate Networks (LGNs) compute through compositions of Boolean operations, yet existing LGNs do not reliably benefit from increased depth. We identify two causes: optimization collapse and topology-induced degradation of output-specific credit that persists even after skip-biased initialization and straight-through estimation stabilize training.

We introduce Input-Anchored Logic Gate Networks (IALGNs), in which each gate combines a private hidden spine with a direct input anchor. This topology prevents output-path merging while retaining input access at every layer. Credit diagnostics show that random wiring dilutes or conflicts output-specific gradients, whereas IALGN maintains usable and coherent credit. Random-$k_x$ relaxation improves anchor selection without relaxing the spine.

Across MNIST, CIFAR-10, and CIFAR-100, IALGN exhibits consistent fixed-width depth--accuracy scaling up to 150 layers, while alternative topologies saturate or degrade. Linear probes, topology ablations, and operation-aware analysis show that trained IALGNs preserve private states and apply sparse anchor-conditioned updates. These results indicate that scalable LGN depth requires both stable optimization and credit-preserving information access.
\end{abstract}


\section{Introduction}

Logic Gate Networks (LGNs) are neural architectures that implement computation through compositions of Boolean operations, closely resembling classical Boolean circuits. By replacing conventional real-valued neurons with logic gates, LGNs provide a structured and interpretable computational model with direct relevance to efficient discrete inference.

From the perspective of Boolean circuit theory, depth is a fundamental computational resource. Increasing circuit depth can strictly enlarge representational power, and broad classes of functions require exponentially larger circuits when restricted to insufficient depth~\cite{BC_depth_scale}. These results suggest that scaling LGNs in depth should be both theoretically meaningful and practically beneficial.

Constructing an LGN requires determining both the logical operation performed by each gate and the wiring that connects gates across layers. Because this design space is discrete and combinatorial, differentiable LGNs replace discrete operation choices with continuous relaxations that can be optimized by gradient descent~\cite{DDLGN,DSLGN}. Despite substantial progress in gate optimization, however, existing LGNs do not exhibit reliable depth--performance scaling. Most reported architectures remain relatively shallow, and increasing depth frequently causes accuracy to saturate or decline. Consequently, the computational benefits predicted by circuit theory remain largely unrealized in practical LGNs.

\begin{figure*}[ht]
    \centering
    \includegraphics[width=1.\linewidth]{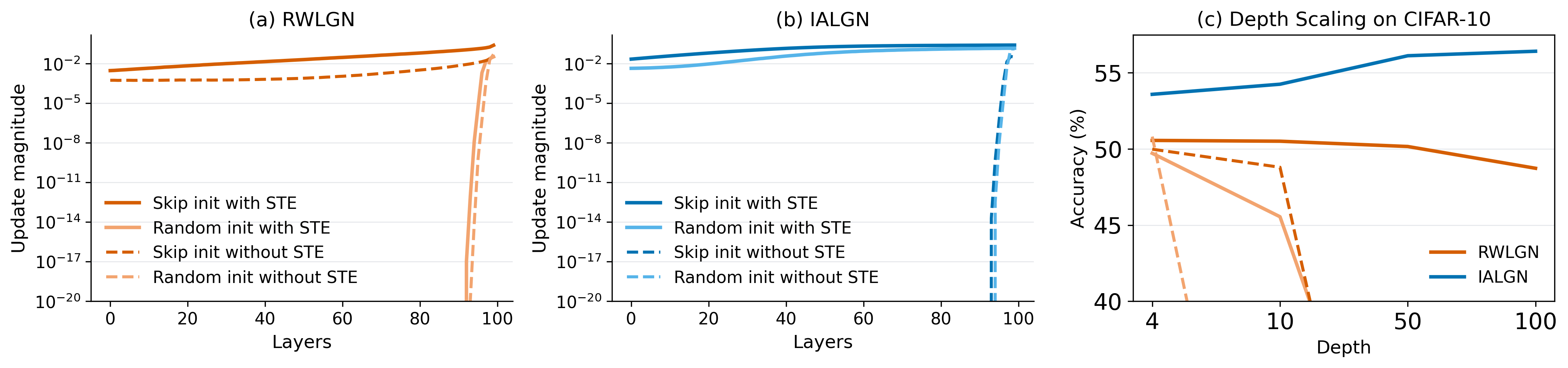}
    \caption{Optimization stability alone is insufficient for depth scalability. (a,b) Layer-wise update magnitudes under different initialization and training settings. Skip-biased initialization and STE substantially improve update propagation across deep networks. (c) Despite stable optimization, RWLGN fails to benefit from increasing depth, whereas IALGN continues to improve, indicating that optimization alone does not explain depth scalability. The IALGN curve in (c) uses Skip init + STE.}
    \label{fig:opt_collapse}
\end{figure*}

This failure is commonly attributed to optimization difficulties. Gradients propagated through many relaxed Boolean operations rapidly vanish, while discrepancies between relaxed training and discrete inference further destabilize learning. Recent advances, including stochastic relaxations, straight-through estimators (STE), and skip-biased initialization, substantially improve trainability and stabilize optimization in deep LGNs~\cite{DSLGN,CDLGN,GAPLGN,LDLGN}. Indeed, under these techniques, layer-wise parameter updates remain large throughout depth in both randomly wired LGNs and our proposed model (Fig.~\ref{fig:opt_collapse}a,b), indicating that optimization collapse is largely mitigated. Nevertheless, stable optimization alone does not translate into depth scalability: even under the strongest optimization protocol, randomly wired LGNs obtain little benefit from increasing depth, whereas our model continues to improve (Fig.~\ref{fig:opt_collapse}c).

We argue that depth scalability requires satisfying two distinct conditions. First, deeper layers must remain trainable (\emph{optimization}). Second, the topology must preserve usable output-specific learning signals for the computations performed at those layers (\emph{credit assignment}). Existing optimization techniques largely address the former but leave the latter unchanged.

To address this limitation, we propose the \textbf{Input-Anchored Logic Gate Network} (IALGN). Each gate combines one feature propagated through a fixed hidden computational spine with one feature directly sampled from the original input. This simple modification allows each layer to preserve or refine accumulated computation using directly available input features. We further show that the resulting topology prevents output-path merging and preserves output-specific credit assignment, separating
the architectural quality of the learning signal from the trainability provided by the optimizer.

Empirically, only IALGN exhibits systematic fixed-width depth scaling across all three datasets. Probe, topology, and operation analyses link this gain to private-state preservation and sparse input-conditioned refinement.

Our contributions are summarized as follows:

\begin{itemize}
    \item We distinguish two obstacles to depth scalability in LGNs: optimization collapse and topology-induced degradation of output-specific credit. We show that resolving aggregate optimization alone does not make additional depth useful.

    \item We propose IALGN, a simple input-anchored topology with random-$k_x$ anchor relaxation. Analyses of path sharing, descendant-conditioned gradient coverage, and gradient purity show that its private input-anchored chains prevent topology-induced degradation of output-specific credit.

    \item We demonstrate consistent fixed-width depth scaling on MNIST, CIFAR-10, and CIFAR-100. Linear probes, topology ablations, and operation-aware depth analysis associate the gains with progressively richer representations and selective refinement of private hidden states, rather than generic wiring flexibility or raw path length.
\end{itemize}

\section{Related Work}

\paragraph{Differentiable Logic Gate Networks.} Differentiable Logic Gate Networks (DLGNs) enable gradient-based learning of Boolean circuits by relaxing the discrete selection of the 16 two-input Boolean functions into a differentiable optimization problem \cite{DDLGN}. The original formulation employs randomly wired feedforward connections together with skip-biased initialization, establishing the foundation for subsequent LGN research.

Recent work has improved DLGNs along several largely orthogonal directions. Convolutional DLGNs extend logic-gate computation to spatial vision tasks through structured logic kernels \cite{CDLGN}. Other studies improve optimization by reducing the discretization gap or introducing more effective gate parameterizations, including stochastic relaxations, straight-through estimators, and lightweight gate representations \cite{DSLGN,GAPLGN,LDLGN}. More recently, recurrent parameter sharing \cite{RDDLGN}, differentiable connectivity optimization \cite{EdgeRelaxLGN}, and hardware-oriented connectivity learning \cite{Fojcik_2026_CVPR} have further broadened the design space of DLGNs. Unlike our work, these approaches primarily modify gate parameterization, optimization, or connectivity learning. They do not isolate whether a fixed, credit-preserving input-access pattern itself enables depth scaling.

\paragraph{Topology and Information Flow in Deep Networks.}
In conventional deep neural networks, architectural connectivity plays a central role in making depth effective. Residual and dense connections improve information propagation by creating short paths between early and late representations, allowing deep networks to progressively refine features rather than repeatedly overwrite them \cite{residual_shallow,topo_grad}. Applying these principles to LGNs is non-trivial because logic gates operate under strict two-input constraints and saturating Boolean semantics, making standard additive skip connections inapplicable. Instead of learning arbitrary hidden connectivity, we introduce a simple topology in which every gate preserves a hidden-layer spine while retaining direct access to the original input. Our approach is therefore complementary to previous LGN research: rather than improving individual gates or their optimization, we investigate whether modifying the information-access pattern itself enables scalable depth.

\begin{figure*}[t]
    \centering
    \includegraphics[width=1.\linewidth]{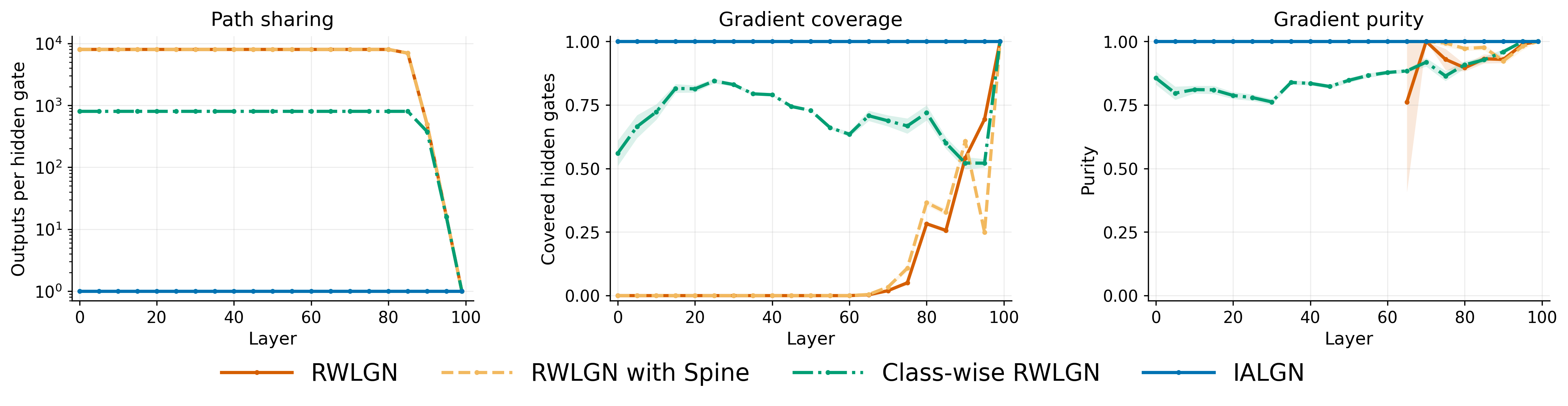}
    \caption{Output-path sharing and output-specific credit assignment in 100-layer CIFAR-10 LGNs. \textbf{Left:} Random wiring rapidly merges output paths; class-wise separation removes only inter-class sharing, whereas IALGN preserves one private lineage per output. \textbf{Middle:} Descendant-conditioned gradient coverage. RWLGN and RWLGN with Spine lose usable credit in early layers. \textbf{Right:} Gradient purity among active gates. Class-wise separation restores coverage but retains interference, whereas IALGN maintains full coverage and unit purity. Diagnostics are measured immediately after skip-biased initialization, before any parameter update, with STE used for gradient computation. Sampling details are provided in the supplement.}
    \label{fig:motif}
\end{figure*}

\section{Why Deep DLGNs Do Not Scale}\label{sec:opt_collapse}
\subsection{Optimization Stability Is Insufficient}
Optimization collapse is a well-known obstacle to training deep Logic Gate Networks (LGNs). Because gradients must propagate through many relaxed Boolean operations, the
learning signal reaching early layers rapidly diminishes as depth increases, often preventing deep models from converging. Recent studies have substantially alleviated this problem through skip-biased initialization and straight-through estimators (STE), enabling stable optimization of considerably deeper LGNs~\cite{DDLGN,DSLGN,GAPLGN,LDLGN}. This naturally raises an important question: if optimization can already be stabilized, why do randomly wired LGNs still fail to benefit from increasing depth?

To separate optimization from architectural effects, we compare randomly wired LGNs (RWLGNs) and the proposed IALGN under four representative training settings: random
initialization, skip-biased initialization, STE, and their combination. Following~\cite{DDLGN}, we measure layer-wise parameter update magnitudes rather than raw gradients, since adaptive optimizers internally rescale gradients before applying parameter updates. Larger update magnitudes therefore indicate that a layer participates more actively in optimization.

Figure~\ref{fig:opt_collapse}(a,b) shows that modern optimization techniques substantially alleviate optimization collapse. Skip-biased initialization greatly improves optimization in RWLGNs, while STE provides particularly strong stabilization in IALGNs. Under the strongest optimization setting (Skip init + STE), parameter updates remain consistently large throughout nearly all layers in both architectures, indicating that optimization collapse is largely mitigated.

Together, Figure~\ref{fig:opt_collapse} shows that stable parameter updates are necessary but insufficient for scalable depth: RWLGN accuracy saturates or degrades, whereas IALGN continues to improve under the same optimization protocol. We therefore examine whether random inter-layer wiring degrades the output-specific credit assigned to hidden computations.

\begin{figure*}[t]
    \centering
    \includegraphics[width=1.\textwidth]{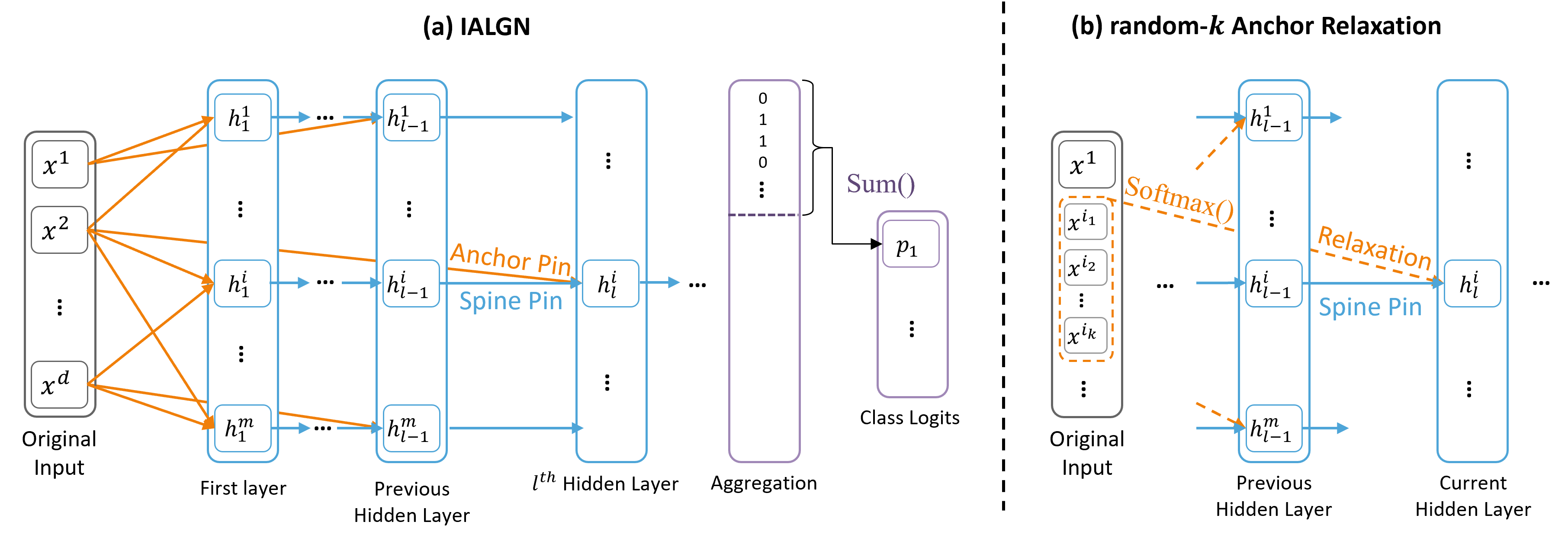}
    \caption{Input-Anchored Logic Gate Networks and random-$k_x$ anchor relaxation. (a) IALGN replaces the fully random inter-layer wiring of standard DLGNs with a two-pin structure. The \emph{spine pin} receives the corresponding hidden feature from the previous layer, preserving a private layer-to-layer computational path. The \emph{anchor pin} receives a feature directly from the original input, making the original input directly available at every layer. (b) Random-$k_x$ anchor relaxation samples $k_x$ candidate input anchors for each gate and learns a relaxed selection among them, improving anchor flexibility without searching over all possible input connections. The standard IALGN retains a fixed identity spine, denoted by $k_h=1$.}
    \label{fig:IALGN}
\end{figure*}

\subsection{Topology-Induced Credit Degradation}

Stable parameter updates do not necessarily imply coherent credit assignment. Let $\mathcal{D}_{\ell,i}$ denote the set of final-layer output gates structurally reachable from hidden gate $i$ at layer $\ell$. We measure \emph{path sharing} by $|\mathcal{D}_{\ell,i}|$. For each final output gate $o$ with activation $z_o$, we define its contribution to the gradient of gate parameters $\theta_{\ell,i}$ as
\[
\mathbf{g}_{o\rightarrow\ell,i}
=
\nabla_{\theta_{\ell,i}}
\left[
\operatorname{stopgrad}
\left(
\frac{\partial\mathcal{L}}{\partial z_o}
\right)z_o
\right].
\]
Summing these contributions over all final output gates
recovers the ordinary loss gradient.

Let
\[
\mathcal{A}_{\ell,i}
=
\left\{
o\in\mathcal{D}_{\ell,i}:
\|\mathbf{g}_{o\rightarrow\ell,i}\|>t_{\ell,i}
\right\}
\]
denote the active output-credit sources, where
$t_{\ell,i}$ is a relative gradient threshold. Gradient
coverage is the fraction of sampled hidden gates with
$\mathcal{A}_{\ell,i}\neq\emptyset$, conditioned on having
at least one sampled structural descendant. Gradient purity is
\[ 
P_{\ell,i}
=
\frac{
\left\|
\sum_{o\in\mathcal{A}_{\ell,i}}
\mathbf{g}_{o\rightarrow\ell,i}
\right\|
}{
\sum_{o\in\mathcal{A}_{\ell,i}}
\|\mathbf{g}_{o\rightarrow\ell,i}\|+\epsilon
}.
\]
A purity of one indicates either a single active credit source or perfectly aligned contributions, whereas lower values indicate directional cancellation. We refer to the absence of active descendant contributions as \emph{credit dilution}, and to cancellation among active contributions as \emph{credit interference}. To avoid architecture-dependent sampling bias, we use the same sampled gate indices for the hidden and output layers, and compute coverage only over hidden gates with at least one structurally reachable sampled output. Additional sampling and threshold details are provided in the supplement.

We compare RWLGN with two controlled variants. In \emph{RWLGN with Spine}, the first pin follows a fixed one-to-one hidden path while the second pin remains randomly connected to the previous hidden layer. In \emph{Class-wise RWLGN}, the same total width is partitioned into class-specific groups, and hidden wiring is restricted within each group.

Figure~\ref{fig:motif} shows that random wiring rapidly merges output paths, such that early RWLGN gates become ancestors of nearly the entire output layer. Introducing a fixed spine does not prevent this entanglement because the second hidden pin reconnects otherwise distinct computational paths. Class-wise separation removes inter-class connections but leaves substantial within-class sharing.

These structural differences are reflected in the credit diagnostics. RWLGN and RWLGN with Spine exhibit severe credit dilution in early layers, whereas class-wise separation restores coverage but retains credit interference among the surviving contributions. In contrast, IALGN preserves one private lineage per output and maintains full sampled coverage with unit purity throughout depth. Thus, random hidden wiring degrades output-specific credit even when aggregate parameter updates remain non-negligible.

\section{Input-Anchored Logic Gate Networks}
\label{sec:ialgn}

\subsection{Design Principle}

The credit-assignment analysis suggests that scalable logical depth requires a stable, output-specific computational lineage. We therefore reserve one gate input pin for a one-to-one hidden path across layers. We instantiate this lineage as an identity \emph{spine}, although any fixed one-to-one permutation would provide the same isolation principle.

A private spine alone is insufficient because a Boolean gate has two input pins. Connecting the second pin to another hidden unit reintroduces cross-chain path merging, as observed in the RWLGN with spine baseline. Feeding the same hidden state to both pins preserves isolation but reduces the gate to a unary Boolean transformation and introduces no new task information. We therefore connect the second pin directly to an original input bit. The input provides a depth-independent and task-relevant source that can enrich each private chain without routing information through another hidden computation.

This yields the Input-Anchored Logic Gate Network (IALGN), in which each layer updates an output-specific hidden state while retaining direct access to the original input. Input anchoring is therefore not an arbitrary skip connection, but a simple way to enrich private computational lineages without reintroducing cross-chain routing.

\begin{figure*}[t]
    \centering
    \includegraphics[width=1.\textwidth]{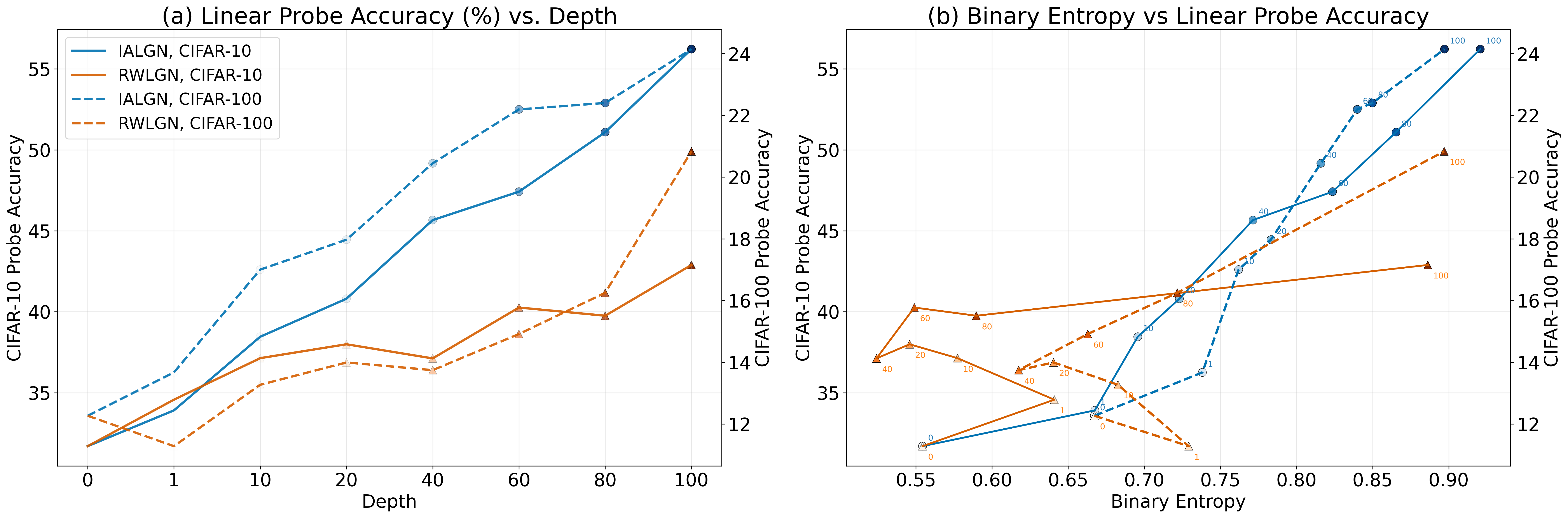}
    \caption{Progressive representation learning enabled by input anchoring. (a) Layer-wise linear-probe accuracy on frozen hidden representations. The probe at depth 0 corresponds to the raw binarized input before any logic-layer computation. Unlike randomly wired DLGNs, IALGN exhibits progressively improving probe accuracy as depth increases on both CIFAR-10 (solid) and CIFAR-100 (dashed), indicating that deeper layers produce increasingly informative representations. (b) Probe accuracy versus average binary activation entropy. Although entropy provides only a supporting diagnostic of hidden-state diversity, IALGN consistently achieves higher probe accuracy across comparable entropy levels, suggesting that its representational improvement cannot be explained solely by activation statistics.
    }
    \label{fig:rep}
\end{figure*}

\subsection{Input-Anchored Topology}

Figure~\ref{fig:IALGN} illustrates the resulting two-pin topology and its random-$k_x$ anchor relaxation. Let $x\in\{0,1\}^{d_{\mathrm{in}}}$ denote the binarized input and $h_\ell\in\{0,1\}^{W}$ the hidden representation at layer $\ell$.

In a conventional randomly wired DLGN, each gate receives both of its inputs from the previous hidden layer,

\begin{equation}
h_\ell^{i}
=
g_\ell^{i}
\left(
h_{\ell-1}^{a_{\ell,i}},
h_{\ell-1}^{b_{\ell,i}}
\right),
\end{equation}
where $g_\ell^{i}$ denotes one of the sixteen Boolean operations and $a_{\ell,i},b_{\ell,i}$ are randomly sampled hidden indices.

IALGN modifies only the source of one gate input,

\begin{equation}
h_\ell^{i}
=
g_\ell^{i}
\left(
h_{\ell-1}^{i},
x^{b_{\ell,i}}
\right).
\end{equation}
The first input $h_{\ell-1}^{i}$, referred to as the \emph{spine pin}, preserves a layer-to-layer computational path through the network. The second input $x^{b_{\ell,i}}$, referred to as the \emph{anchor pin}, connects every layer directly to the original input. For the first layer, both inputs are drawn from the binarized input.

Importantly, IALGN changes only the source of one gate pin; the Boolean gate family, differentiable relaxation, objective, and readout remain identical to those of standard DLGNs.

\paragraph{Random-$k_x$ Anchor Relaxation.}
A fixed input anchor may be restrictive when the input dimension is large. We therefore sample $k_x$ candidate input anchors for each gate and learn a relaxed selection among them. The case $k_x=1$ corresponds to a fixed anchor. For later ablations, we similarly use $k_h$ to denote the number of candidate hidden features for the spine pin; $k_h=1$ denotes the standard fixed same-index spine.

\paragraph{Training and Inference.} IALGN uses the same differentiable relaxation over the sixteen Boolean operations as standard DLGNs, together with learnable anchor-selection logits. Unless otherwise specified, all experiments use skip-biased initialization and hard-forward STEs for both gate and anchor selection: the forward pass uses argmax choices, while the corresponding soft relaxations provide surrogate gradients. Inference therefore evaluates the same discretized two-input Boolean circuit used during training.

\section{Progressive Representation Learning}\label{sec:5}

Section~\ref{sec:ialgn} introduced an input-anchored topology that preserves private computational lineages while giving every layer direct access to the original input. We next examine whether this architecture produces increasingly useful hidden
representations as depth increases.

\subsection{Linear Probe Analysis} To measure the information contained in intermediate representations, we train an independent linear classifier on the frozen hidden representation produced by each layer. Since the probe itself introduces no additional nonlinear computation, its accuracy directly reflects the linear separability of the learned representation.

Figure~\ref{fig:rep}(a) reports the layer-wise probe accuracy on CIFAR-10 and CIFAR-100. Randomly wired DLGNs exhibit little improvement beyond the early layers, and probe accuracy gradually deteriorates as depth increases. This behavior suggests that repeatedly recombining transformed hidden features provides limited representational benefit.

In contrast, IALGN exhibits a steady improvement in probe accuracy throughout depth on both datasets. As additional layers preserve the accumulated hidden computation while
retaining access to direct input anchors, intermediate representations become progressively easier to decode. This observation is consistent with the credit-assignment analysis in Section~\ref{sec:opt_collapse}: cleaner output-specific credit is associated with progressively more decodable representations.

\paragraph{Activation Entropy.}
Figure~\ref{fig:rep}(b) compares probe accuracy with average binary activation entropy.
IALGN remains more linearly separable at comparable entropy levels, indicating
that its improvement cannot be explained solely by activation statistics.

\begin{table*}[t]
\centering
\caption{Fixed-width depth scaling results. Within each row, RWLGN and IALGN use identical width and depth and therefore the same number of deployed logic gates. Results in Tables~1 and~2 are averaged over five seeds; figure-specific aggregation is stated in the corresponding captions.. LDLGN reimpl. denotes our reimplementation under the same no-augmentation protocol and should be interpreted as a within-paper reference, not as a reproduction of the original reported results. The best result in each row is shown in bold.}
\begin{tabular}{llll|ccc}
\toprule
Dataset   & Width & Depth & Gates & RWLGN & LDLGN reimpl. & IALGN \\
\midrule
CIFAR-10  & 12k   & 4     & 48k   & $50.56 \pm 0.58$ & $50.22 \pm 0.18$ & \textbf{53.59 $\pm$ 0.04} \\
CIFAR-10  & 12k   & 10    & 120k  & $50.51 \pm 0.63$ & $50.87 \pm 0.36$ & \textbf{54.25 $\pm$ 0.06} \\
CIFAR-10  & 12k   & 50    & 600k  & $50.16 \pm 0.33$ & $45.40 \pm 0.38$ & \textbf{56.12 $\pm$ 0.05} \\
CIFAR-10  & 12k   & 100   & 1.2M  & $48.73 \pm 0.50$ & $38.85 \pm 0.87$ & \textbf{56.41 $\pm$ 0.17} \\
CIFAR-10  & 12k   & 150   & 1.8M  & $47.91 \pm 0.09$ & $22.72 \pm 1.31$ & \textbf{56.50 $\pm$ 0.11} \\
\midrule
CIFAR-10  & 128k  & 4     & 512k    & $57.01 \pm 0.42$ & $54.01 \pm 0.15$ & \textbf{57.31 $\pm$ 0.47} \\
CIFAR-10  & 128k  & 12    & 1.536M  & $56.12 \pm 0.20$ & $54.31 \pm 0.21$ & \textbf{58.72 $\pm$ 0.58} \\
CIFAR-10  & 128k  & 20    & 2.560M  & $55.97 \pm 0.09$ & $53.87 \pm 0.39$ & \textbf{59.23 $\pm$ 0.27} \\
CIFAR-10  & 128k  & 80    & 10.24M  & $55.66 \pm 0.17$ & $52.05 \pm 0.11$ & \textbf{59.61 $\pm$ 0.19} \\
\midrule
CIFAR-100 & 128k  & 4     & 512k   & $25.77 \pm 0.25$ & $26.43 \pm 0.19$ & \textbf{28.84 $\pm$ 0.25} \\
CIFAR-100 & 128k  & 12    & 1.536M & $26.65 \pm 0.27$ & $27.00 \pm 0.08$ & \textbf{29.46 $\pm$ 0.40} \\
CIFAR-100 & 128k  & 20    & 2.560M & $26.31 \pm 0.14$ & $27.04 \pm 0.16$ & \textbf{30.29 $\pm$ 0.26} \\
CIFAR-100 & 128k  & 80    & 10.24M & $26.19 \pm 0.27$ & $25.16 \pm 0.21$ & \textbf{30.60 $\pm$ 0.06} \\
\midrule
MNIST     & 8k    & 6     & 48k   & $97.29 \pm 0.06$ & $97.54 \pm 0.03$ & \textbf{98.22 $\pm$ 0.02} \\
MNIST     & 8k    & 100   & 800k  & $97.12 \pm 0.06$ & $94.31 \pm 0.08$ & \textbf{98.33 $\pm$ 0.07} \\
\bottomrule

\end{tabular}
\label{tab:depth_scaling}
\end{table*}

\begin{figure}[!ht]
    \centering
    \includegraphics[width=1.\linewidth]{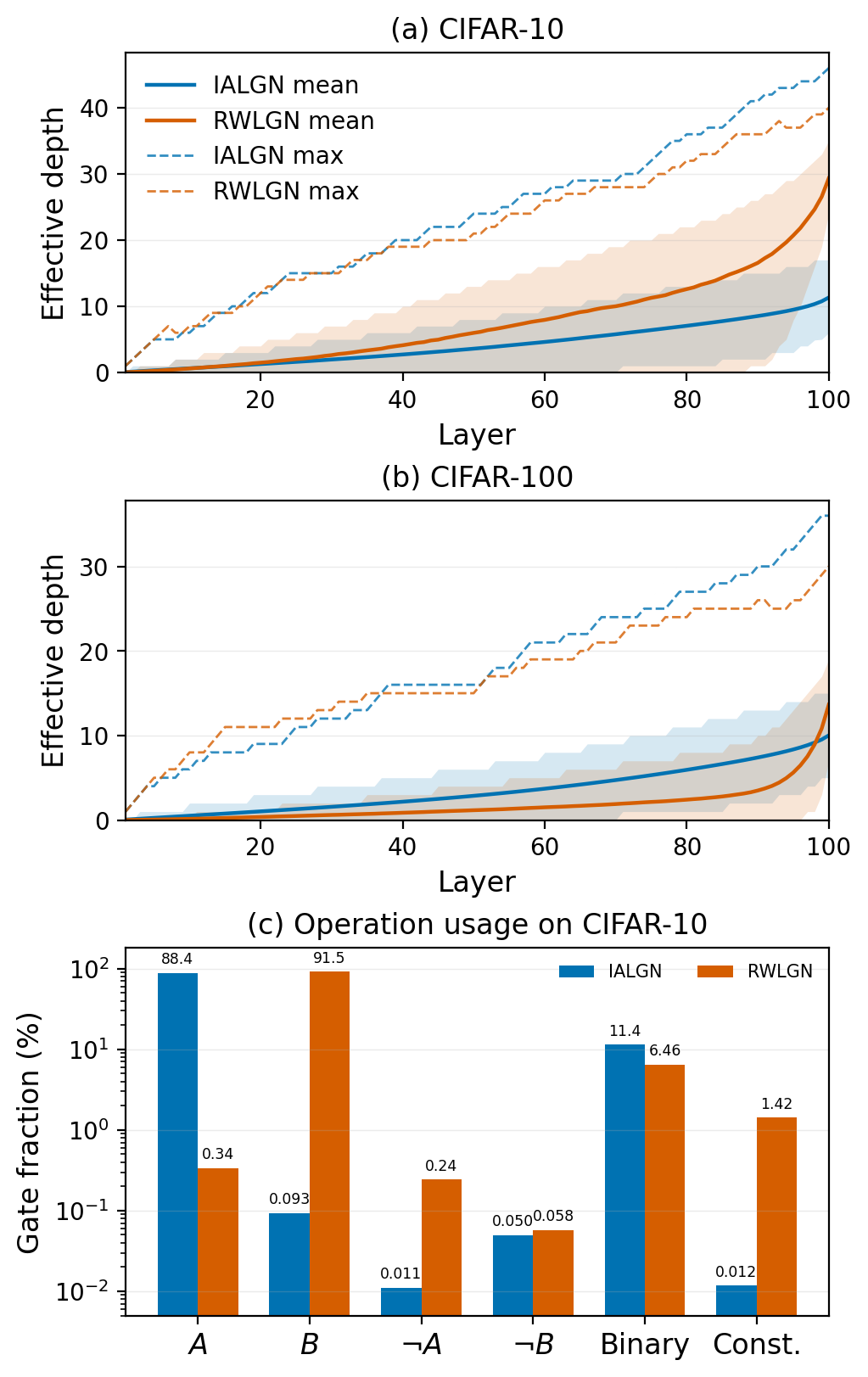}
    \caption{Selective depth utilization in trained 100-layer networks. (a,b) Solid and dashed lines denote mean and maximum operation-aware effective depth, respectively; shaded regions show the 10th--90th percentiles over gates. At the final layer, RWLGN has larger mean depth, whereas IALGN reaches larger maxima on both datasets. (c) CIFAR-10 operation frequencies, excluding the first layer. For IALGN, $A$ and $B$ denote the spine and anchor pins; for RWLGN, both pins receive hidden predecessors. IALGN predominantly preserves its spine and applies sparse binary updates, whereas RWLGN mainly passes through randomly wired hidden states.}
    \label{fig:depth_analysis}
\end{figure}

\section{Experiments}\label{sec:experiments}

We evaluate IALGN from three complementary perspectives:
(i) whether increasing depth improves performance at fixed width,
(ii) whether the gains arise from input anchoring rather than generic wiring
flexibility, and (iii) how trained circuits use nominal depth through
state preservation and anchor-conditioned updates. Additional sensitivity
analyses on input binarization and the anchor candidate size $k_x$ are
provided in the supplementary material.

\subsection{Experimental Setup}

We evaluate RWLGNs and IALGNs on MNIST, CIFAR-10, and CIFAR-100 without data augmentation. Each normalized input $v\in[0,1]$ is thermometer-encoded as $\{\mathbf{1}[v>j/n_b]\}_{j=1}^{n_b-1}$, with $n_b=4$ for MNIST and $n_b=16$ for CIFAR. Unless otherwise stated, IALGN uses $k_x=32$ with a fixed spine ($k_h=1$), and all results are averaged over five seeds.

At fixed width, varying depth changes the deployed gate count linearly. This experiment therefore tests whether added layers yield useful computation, rather than superiority under an identical total-gate budget. RWLGN and IALGN share the same operations, readout, objective, initialization, and STE, differing primarily in wiring topology. Raw-gradient and random-initialization variants are analyzed in Section~\ref{sec:opt_collapse} and the supplementary material. We also report an LDLGN reimplementation under the same no-augmentation protocol as a controlled reference, since we could not identify an official implementation at the time of submission.

\subsection{Depth Scaling}

Table~\ref{tab:depth_scaling} summarizes the classification accuracy on MNIST, CIFAR-10, and CIFAR-100. Across all three benchmarks, RWLGNs fail to obtain consistent
benefits from increasing depth and often degrade, with severe performance loss occurring in some deep configurations. In contrast, IALGNs consistently benefit from additional depth and progressively improve classification accuracy.

The improvement is particularly pronounced on more challenging datasets such as CIFAR-10 and CIFAR-100, where deeper IALGNs continue to gain accuracy while conventional DLGNs saturate or deteriorate. These results indicate that the proposed topology converts added layers into useful computation rather than representation degradation, consistent with the credit-assignment and representation analyses in Sections 3 and 5. The following ablation investigates whether these gains arise from the proposed information-access pattern itself rather than simply from increasing wiring flexibility.

\subsection{Hidden-Wiring $k_h$ Ablation}\label{sec:ablation2}

\begin{table}[t]
\centering
\small
\caption{Hidden-$k_h$ relaxation is added on top of an input-anchored IALGN with $k_x=32$ to evaluate whether additional hidden-to-hidden flexibility benefits deep models.}
\begin{tabular}{lcc}
\toprule
Topology(D=4) & Relaxed source & Accuracy \\
\midrule
RWLGN  & none & $50.56 \pm 0.58$ \\
LDLGN reimpl. & none & $50.22 \pm 0.18$ \\
\midrule
IALGN (${k_x}=32, {k_h}=1$)  & input & $53.59 \pm 0.04$ \\
IALGN (${k_x}=32, {k_h}=8$)  & input, hidden & $53.91 \pm 0.27$ \\
IALGN (${k_x}=32, {k_h}=16$) & input, hidden & \textbf{54.11 $\pm$ 0.29} \\
IALGN (${k_x}=32, {k_h}=32$) & input, hidden & $53.73 \pm 0.11$ \\
\midrule
Topology(D=50) & Relaxed source & Accuracy \\
\midrule
RWLGN  & none & $50.16 \pm 0.33$ \\
LDLGN reimpl. & none & $45.40 \pm 0.38$ \\
\midrule
IALGN (${k_x}=32, {k_h}=1$)  & input & \textbf{56.12 $\pm$ 0.05} \\
IALGN (${k_x}=32, {k_h}=8$)  & input, hidden & $55.66 \pm 0.12$ \\
IALGN (${k_x}=32, {k_h}=16$) & input, hidden & $55.23 \pm 0.23$ \\
IALGN (${k_x}=32, {k_h}=32$) & input, hidden & $55.11 \pm 0.14$ \\
\bottomrule
\end{tabular}
\label{tab:hidden_k}
\end{table}

Starting from the primary IALGN configuration ($k_x=32,k_h=1$), we relax the spine pin over $k_h\in\{8,16,32\}$ hidden candidates while keeping $k_x$ fixed. This isolates whether generic hidden-wiring flexibility benefits depth scaling.

As shown in Table~\ref{tab:hidden_k}, hidden-$k_h$ relaxation provides a modest benefit at $D=4$ but consistently degrades performance at $D=50$. Thus, the gain does not arise from generic wiring flexibility; preserving the fixed spine becomes increasingly important with depth.

\subsection{Selective Depth Utilization}\label{sec:ablation3}
Nominal depth need not add one non-trivial operation per layer. We therefore compute operation-aware effective depth after discretization: constants reset depth to zero, pass-through operations inherit the relevant input depth, unary negations add one, and genuinely binary operations add one to the longer input path. Direct input anchors have depth zero; full details are provided in the supplementary material.

Figures~\ref{fig:depth_analysis}(a,b) separate average from extreme path utilization. At the final layer, RWLGN has larger mean effective depth, whereas IALGN reaches larger maximum depths on both datasets. Thus, raw path statistics alone do not explain depth scalability. RWLGN depth is assembled through branching and broadly shared hidden ancestry, whereas IALGN preserves isolated, output-specific spines.

Figure~\ref{fig:depth_analysis}(c) clarifies how IALGN uses these spines. In the analyzed CIFAR-10 circuit, $88.4\%$ of gates select the spine-preserving operation $A$, while $11.4\%$ perform binary anchor-conditioned updates. Fewer than $0.2\%$ discard the accumulated spine through $B$, $\neg B$, or constants. IALGN therefore uses nominal depth as a long-lived private state with sparse refinements. RWLGN also favors pass-through operations, but they follow randomly wired hidden predecessors, consistent with the path sharing and degraded credit observed in Figure~\ref{fig:motif}.

These statistics distinguish nominal layer utilization from functional path length. An IALGN gate may preserve its hidden state across many layers through $A$ without increasing effective depth, while retaining the opportunity for a later anchor-conditioned update. The relatively small mean effective depth therefore reflects selective transformation, not the absence of a depth-wise computational state. This also reconciles Figure~\ref{fig:depth_analysis} with the probe results in Figure~\ref{fig:rep}: useful depth need not maximize the number of realized operations, but must preserve a coherent state on which later operations can act. RWLGN may accumulate longer average paths, yet those paths are assembled through shared hidden ancestry and do not yield the same progressive decodability.

\section{Conclusion}
We showed that depth scalability in LGNs requires solving two distinct problems: optimization collapse and topology-induced degradation of output-specific credit. Even when skip-biased initialization and straight-through estimation stabilize deep randomly wired LGNs, additional layers do not reliably improve their representations or predictive performance. By addressing both requirements, IALGN produces consistent fixed-width accuracy gains with depth across all three datasets, including configurations of up to 150 layers.

IALGN addresses the second limitation by combining a private depth-wise spine with direct input anchors. This topology prevents the rapid path merging observed in randomly wired LGNs and preserves output-specific credit across depth. Layer-wise credit diagnostics show that IALGN maintains near-complete descendant-conditioned gradient coverage and consistently high gradient purity, whereas a hidden spine or class-wise separation alone does not fully resolve the problem. Linear probes, topology ablations, and depth-utilization analysis further associate this cleaner credit structure with more informative representations and selective anchor-conditioned refinement of private hidden states, rather than with raw path length alone.

Our study is limited to feedforward, flattened LGNs and does not establish superiority at every fixed gate budget. Future work includes budget-matched depth--width trade-offs, structured anchors, convolutional variants, and hardware-aware implementations. Scalable depth therefore requires jointly designed optimization and information access.


\bibliography{references}

\clearpage

\appendix

\begin{table}[t]
\centering
\caption{Default hyperparameter configuration used throughout the
experiments unless otherwise specified.}
\label{tab:supp_hparam}
\begin{tabular}{lc}
\toprule
Hyperparameter & Value \\
\midrule
Optimizer & Adam \\
Learning rate & 0.01 \\
Training epochs & 200 \\
Batch size & 100 \\
Initialization & Skip-biased \\
Training method & STE \\
Default MNIST binarization resolution ($n_b$) & 4 \\
Default CIFAR binarization resolution ($n_b$) & 16 \\
Default anchor candidate count ($k_x$) & 32 \\
Default spine candidate count ($k_h$) & 1 \\
\bottomrule
\end{tabular}
\end{table}

\section{Experimental Details}

\subsection{Experimental Setup}

All experiments were conducted on a single NVIDIA RTX A6000 GPU with 48\,GB of memory, running Ubuntu~22.04, Python~3.10, CUDA~12.8, and PyTorch~2.6.0.

Following prior LGN work, all experiments use the Adam optimizer with a learning rate of $0.01$ for $200$ training epochs. Unless otherwise specified, the batch size is fixed to $100$, and skip-biased initialization together with the straight-through estimator (STE) is employed throughout training. To facilitate controlled comparison, these optimization settings are kept identical across the compared methods.

Following prior LGN work, each normalized input value $v\in[0,1]$ is thermometer-encoded as $\{\mathbf{1}[v>j/n_b]\}_{j=1}^{n_b-1}$, with $n_b=4$ for MNIST and $n_b=16$ for CIFAR-10 and CIFAR-100. Unless otherwise stated, IALGN uses random-$k_x$ input anchoring with $k_x=32$, and only the variable under investigation is changed while all other hyperparameters remain fixed.

\subsection{Dataset Preprocessing}
MNIST, CIFAR-10, and CIFAR-100 are evaluated using their standard train/test splits. Unlike the original LDLGN experimental protocol, no random cropping, horizontal flipping, or other data augmentation is used in our controlled comparisons.

\subsection{Randomness and Evaluation}

Unless otherwise stated, all classification results reported in the main paper are averaged over five independent random seeds, and the corresponding standard deviations are reported.

For the linear-probe experiments, we analyze five independently trained logic-network checkpoints. For each checkpoint, the probe is trained from five independent initializations; reported accuracies are averaged first over probe initializations and then over network seeds.

\subsection{Code Availability}

The complete implementation and training scripts will be released publicly upon acceptance to facilitate reproducibility.

\begin{table}[t]
\centering
\caption{Dataset-specific logit scaling ($\tau$).}
\label{tab:supp_tau}
\begin{tabular}{lc}
\toprule
Experiment & $\tau$ \\
\midrule
MNIST & 10 \\
CIFAR-10 ($W=12$k) & 30 \\
CIFAR-10 ($W=128$k) & 100 \\
CIFAR-100 ($W=128$k) & 30 \\
\bottomrule
\end{tabular}
\end{table}

\section{Credit-Assignment Diagnostic Details}
\label{sec:supp_credit}

\paragraph{Experimental setting.}
The diagnostics are measured immediately after initialization, before any parameter updates are applied. The credit-assignment diagnostics in Figure~2 of the main paper are computed on 100-layer CIFAR-10 models with hidden width $W=12{,}000$, using skip-biased operation initialization and the straight-through estimator for gradient computation. For each architecture, we sample $300$ gate indices approximately uniformly across the contiguous class-specific output groups and average the gradient-based statistics over $20$ mini-batches. The shaded regions indicate one standard deviation across mini-batches.

\paragraph{Paired hidden--output sampling.}
A global random sample of output gates would systematically underestimate gradient coverage in private-chain architectures: an IALGN hidden gate has only one structurally corresponding final output, whereas a randomly wired hidden gate can reach many outputs. To avoid this architecture-dependent sampling bias, we use the same sampled gate indices at every analyzed hidden layer and at the final output layer. Coverage is computed only over sampled hidden gates having at least one structurally reachable sampled output. This guarantees that the corresponding output of each sampled IALGN chain is included while retaining an identical sampled index set across architectures.

\paragraph{Structural path sharing.}
For hidden gate $i$ at layer $\ell$, let $\mathcal{D}_{\ell,i}$ denote the set of all final-layer output gates structurally reachable through hidden-to-hidden connections. Path sharing is measured as $|\mathcal{D}_{\ell,i}|$ and averaged over the sampled hidden gates. The traversal follows all registered hidden pins but excludes direct input-anchor pins. It therefore measures potential topological sharing independently of the Boolean operation currently selected at each gate. For the widths used in Figure~2, the descendant sets are computed exactly.

Let $\mathcal{O}_{s}$ denote the sampled final-output indices, and define
\[
\widetilde{\mathcal{D}}_{\ell,i}
=
\mathcal{D}_{\ell,i}
\cap
\mathcal{O}_{s}
\]
as the structurally reachable sampled outputs of hidden gate $i$. Path sharing is computed using the full descendant set $\mathcal{D}_{\ell,i}$, whereas the gradient-based diagnostics use $\widetilde{\mathcal{D}}_{\ell,i}$.

\paragraph{Output-specific gradient contributions.}
For final output activation $z_o$ and operation parameters
$\theta_{\ell,i}$ of hidden gate $i$, we compute
\[
\mathbf{g}_{o\rightarrow\ell,i}
=
\nabla_{\theta_{\ell,i}}
\left[
\operatorname{stopgrad}
\left(
\frac{\partial\mathcal{L}}{\partial z_o}
\right)
z_o
\right].
\]
This provides an exact first-order decomposition of the ordinary loss gradient:
\[
\nabla_{\theta_{\ell,i}}\mathcal{L}
=
\sum_o
\mathbf{g}_{o\rightarrow\ell,i}.
\]
Only contributions from structurally reachable sampled outputs $o\in\widetilde{\mathcal{D}}_{\ell,i}$ are used in the subsequent diagnostics.

\paragraph{Active credit and gradient coverage.}
For each sampled hidden gate, an output-specific contribution is considered active when
\[
\left\|
\mathbf{g}_{o\rightarrow\ell,i}
\right\|
>
t_{\ell,i},
\]
where
\[
t_{\ell,i}
=
\max\left(
10^{-12},
10^{-5}
\max_{o'\in\widetilde{\mathcal{D}}_{\ell,i}}
\left\|
\mathbf{g}_{o'\rightarrow\ell,i}
\right\|
\right).
\]
We define the active sampled output-credit sources as
\[
\mathcal{A}_{\ell,i}
=
\left\{
o\in\widetilde{\mathcal{D}}_{\ell,i}
:
\left\|
\mathbf{g}_{o\rightarrow\ell,i}
\right\|
>
t_{\ell,i}
\right\}.
\]
Descendant-conditioned gradient coverage is the fraction of sampled hidden gates with $\mathcal{A}_{\ell,i}\neq\emptyset$, conditioned on the gate having at least one structurally reachable sampled output, i.e., $\widetilde{\mathcal{D}}_{\ell,i}\neq\emptyset$. Thus, a zero value indicates that structurally available sampled output paths fail to provide a numerically usable credit signal, rather than that no such path exists.

\paragraph{Gradient purity.}
For each hidden gate receiving active credit, gradient purity is defined as
\[
P_{\ell,i}
=
\frac{
\left\|
\sum_{o\in\mathcal{A}_{\ell,i}}
\mathbf{g}_{o\rightarrow\ell,i}
\right\|
}{
\sum_{o\in\mathcal{A}_{\ell,i}}
\left\|
\mathbf{g}_{o\rightarrow\ell,i}
\right\|
+\epsilon
}.
\]
where $\epsilon=10^{-12}$ is used for numerical stability.

Purity is averaged only over hidden gates with nonempty active sets. A value of one occurs when a gate receives a single active output-specific contribution or when all active contributions are perfectly aligned. Lower values indicate directional cancellation among the surviving credit signals. Consequently, purity must be interpreted jointly with coverage: high purity does not imply useful credit assignment when almost no hidden gates receive active credit.

\section{Hard-Forward STE and Deployment Equivalence}

The primary IALGN experiments use straight-through estimators for both input-anchor selection and Boolean-operation selection. For anchor selection, let $\widetilde{x}_{\mathrm{soft}}$ denote the softmax-weighted mixture of the $k_x$ anchor candidates and let $\widetilde{x}_{\mathrm{hard}}$ denote the candidate selected by argmax. The anchor supplied to the gate is
\[
\widetilde{x}_{\mathrm{STE}}
=
\widetilde{x}_{\mathrm{soft}}
+
\operatorname{stopgrad}
\left(
\widetilde{x}_{\mathrm{hard}}
-
\widetilde{x}_{\mathrm{soft}}
\right).
\]
Analogously, if $h_{\mathrm{soft}}$ and $h_{\mathrm{hard}}$ denote the relaxed
and argmax-selected Boolean-gate outputs, respectively, the layer output is
\[
h_{\mathrm{STE}}
=
h_{\mathrm{soft}}
+
\operatorname{stopgrad}
\left(
h_{\mathrm{hard}}-h_{\mathrm{soft}}
\right).
\]
Both estimators therefore use hard choices in the forward pass while propagating surrogate gradients through their soft counterparts. Consequently, the forward computation used during training is exactly the discretized two-input Boolean circuit obtained by argmax selection at inference. The primary STE-trained models therefore introduce no post-hoc continuous--discrete accuracy gap.

\section{Representation-Diagnostic Details}

\paragraph{Linear probes.}
For each analyzed layer, the trained logic network is frozen, and its hard-forward hidden representations are extracted for the training and test sets. An independent linear classifier is trained using Adam for 200 epochs with learning rate 0.01 and batch size 100. We evaluate layers $\ell\in\{0, 1, 10, 20, 40, 60, 80, 100\}$. For each of five independently trained logic-network checkpoints, the probe is trained from five independent initializations. Reported accuracies are first averaged over the five probe runs and then over the five network seeds.

\paragraph{Activation entropy.}
Let $p_{\ell,i}$ be the empirical activation frequency of hidden gate $i$
at layer $\ell$ over the evaluation set. Its binary entropy is
\[
H_{\ell,i}
=
-p_{\ell,i}\log_2 p_{\ell,i}
-(1-p_{\ell,i})\log_2(1-p_{\ell,i}),
\]
where $0\log 0=0$. Layer entropy is the average of $H_{\ell,i}$ over
all hidden gates.

\section{Additional Experimental Results}

\subsection{Initialization and Gradient-Estimator Variants}

\begin{figure}[h]
    \centering
    \includegraphics[width=1.\linewidth]{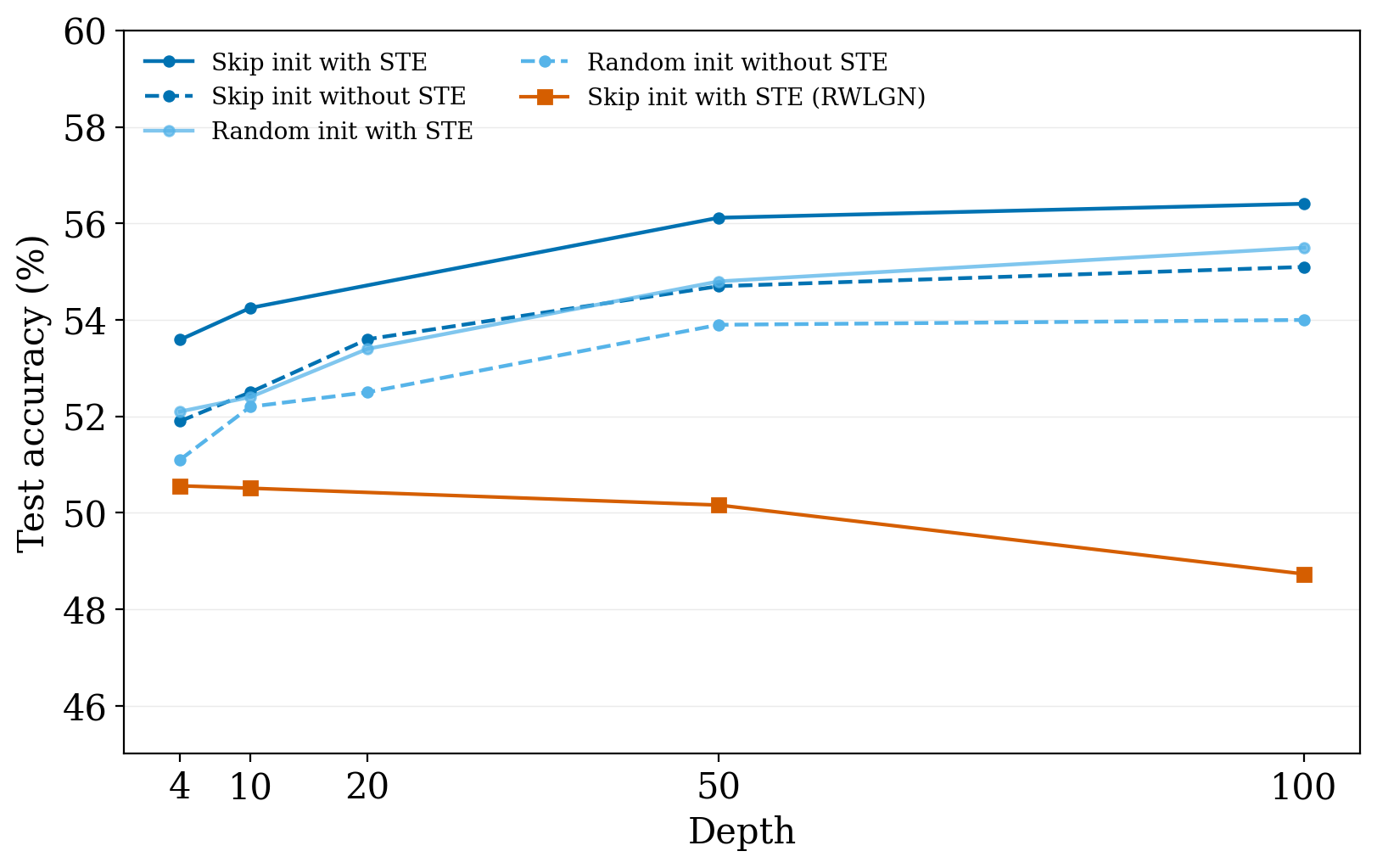}
    \caption{Depth scaling under initialization and gradient-estimator variants on CIFAR-10 with $W=12k$. The four blue curves correspond to IALGN: darker and lighter shades indicate skip-biased and random initialization, respectively, while solid and dashed lines indicate training with and without the hard-forward STE. The orange curve shows RWLGN under the strongest shared protocol, skip-biased initialization with STE. Optimization choices affect the absolute accuracy of IALGN, but its performance generally improves with depth across all four variants. In contrast, RWLGN degrades with depth even under the strongest optimization setting.}
    \label{fig:init_gradient_variants}
\end{figure}

To determine whether the observed depth scaling can be attributed solely to skip-biased initialization or the straight-through estimator, we evaluate IALGN under all four combinations of skip-biased or random initialization and training with or without the hard-forward STE. All experiments use CIFAR-10 with $W=12k$, $n_b=16$, $k_x=32$, and $k_h=1$. We additionally include RWLGN under skip-biased initialization with STE, which is the strongest optimization protocol shared by the two architectures.

Figure~\ref{fig:init_gradient_variants} shows that optimization choices substantially affect the absolute performance of IALGN. Skip-biased initialization combined with STE achieves the strongest results, increasing from $53.6\%$ at $D=4$ to approximately $56.4\%$ at $D=100$. Nevertheless, the qualitative depth-scaling trend is not restricted to this setting. IALGN with random initialization and without STE improves from approximately $51.1\%$ to $54.0\%$, while the variant with random initialization and raw relaxed gradients improves through $D=50$ and remains above its shallow performance at $D=100$. The skip-initialized variant without STE exhibits a similar increasing trend.

In contrast, RWLGN decreases from $50.6\%$ at $D=4$ to $48.7\%$ at $D=100$ despite using skip-biased initialization and STE. These results separate the effects of optimization from those of topology. Improved initialization and gradient estimation raise the attainable accuracy and facilitate deep training, but they do not by themselves determine whether additional layers become useful. The consistent difference in depth trends supports the main conclusion that the input-anchored topology is responsible for converting increased depth into improved predictive performance.

\subsection{Depth Scaling with Fixed Input Anchors}

\begin{figure}[!t]
    \centering
    \includegraphics[width=1.\linewidth]{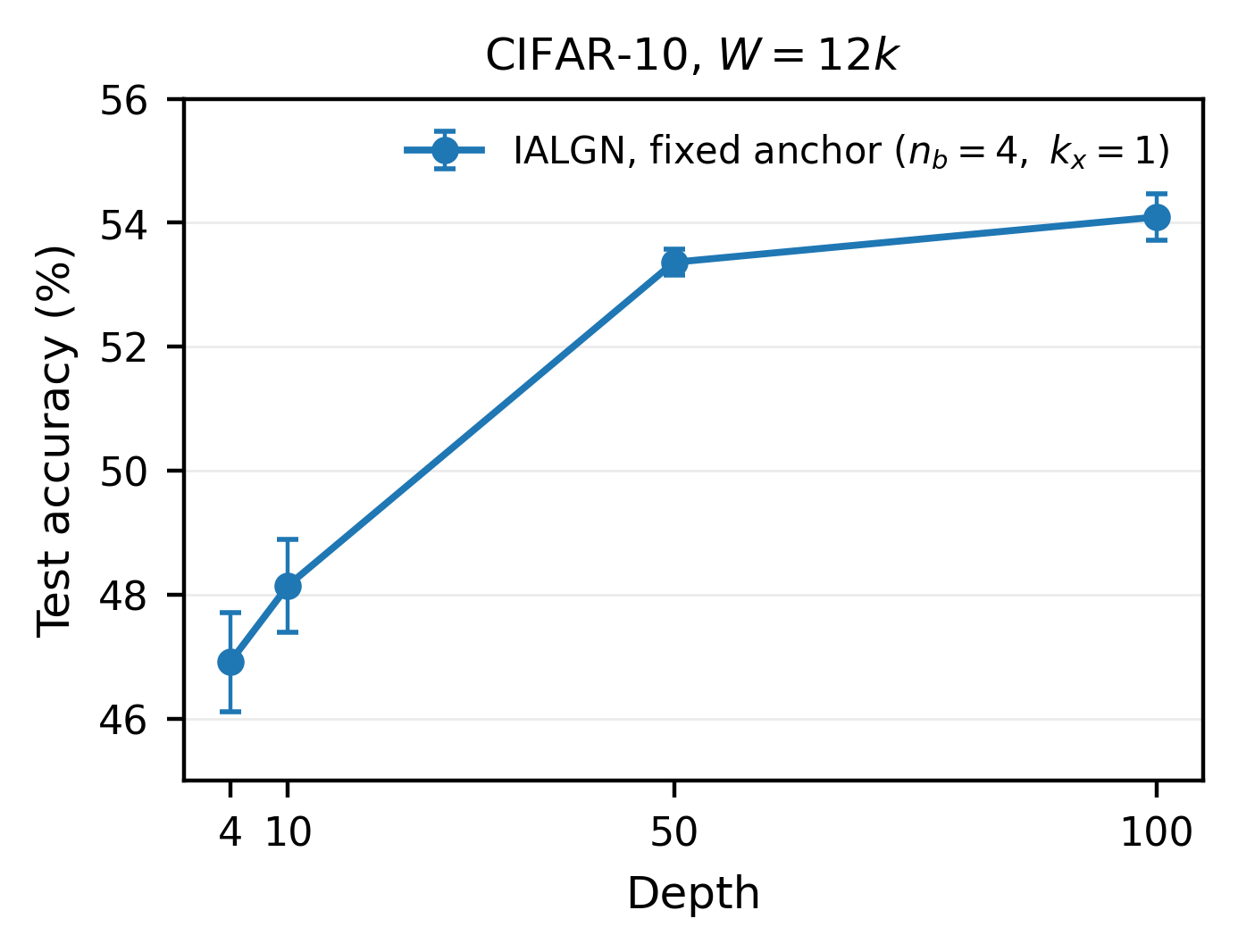}
    \caption{Depth scaling of fixed-anchor IALGN on CIFAR-10 at $W=12k$. The model uses $n_b=4$, $k_x=1$, and $k_h=1$. Despite its low shallow accuracy, performance improves substantially as depth increases, showing that learned random-$k_x$ anchor selection is not required for an input-anchored network to benefit from additional layers. Error bars denote one standard deviation over five independent seeds.}
    \label{fig:fixed_anchor_scaling}
\end{figure}

To test whether direct input anchoring itself is sufficient to alter the depth-scaling trend, we evaluate a fixed-anchor IALGN with $k_x=1$. A single randomly assigned anchor is particularly sensitive to the size of the encoded input space; using $n_b=16$ divides each input value into many candidate bits while providing each gate access to only one of them. Accordingly, this diagnostic uses $n_b=4$ and compares the fixed-anchor IALGN against RWLGN under the same encoding, width, initialization, and
training protocol.

Figure~\ref{fig:fixed_anchor_scaling} shows that the fixed-anchor IALGN improves substantially with depth, despite its low shallow accuracy. This positive trend demonstrates that learned random-$k_x$ anchor selection is not required for an input-anchored topology to convert additional layers into useful computation. Random-$k_x$ relaxation should therefore be viewed as a mechanism for improving anchor-selection flexibility and absolute accuracy, rather than as the sole cause of depth scaling.

For reference, we also show the default IALGN configuration with $n_b=16$ and $k_x=32$. This default configuration attains higher absolute accuracy, although the difference cannot be attributed separately to input resolution or anchor candidate size because both are changed simultaneously. Because both $n_b$ and $k_x$ differ from the fixed-anchor diagnostic, the gap between these two IALGN curves should not be interpreted as an isolated effect of $k_x$. Their interaction is examined separately in Figure~3.

\subsection{Accuracy versus Deployed Gate Count}

The primary depth-scaling experiments hold the hidden width fixed while increasing the number of layers. This isolates whether additional depth can be converted into useful computation, but it also increases the deployed circuit size linearly. We therefore examine test accuracy as a function of the number of deployed logic gates. For a network with width $W$ and depth $D$, the deployed gate count is $W D$. The random-$k_x$ anchor candidates are used only during training and do not affect this count, since each gate retains a single selected anchor at inference.

Figure~\ref{fig:gate_count_accuracy} connects the depth sweeps from Table~1 of the main paper separately within each fixed-width family. On CIFAR-10, IALGN exhibits an increasing accuracy trend in both the $W=12k$ and $W=128k$ regimes. In contrast, increasing the gate count of RWLGN produces little benefit and eventually reduces accuracy, while the LDLGN reimplementation degrades substantially in its deeper configurations. The wider IALGN models achieve higher absolute accuracy, but the positive depth trend is present within both width families.

The comparison is even clearer on CIFAR-100, where all methods use $W=128k$. IALGN improves from $28.84\%$ to $30.60\%$ over the evaluated gate-count range, whereas RWLGN remains approximately saturated and LDLGN declines at the largest depth. Thus, the fixed-width improvements reported in the main paper remain visible when model size is expressed in deployed logic gates.

This analysis does not establish that depth is universally preferable to width under every exactly matched gate budget. The evaluated points represent a finite set of width--depth combinations rather than a complete budget-matched architecture search. Instead, the result shows that IALGN converts increases in deployed circuit size into consistent accuracy gains more effectively than the alternative topologies over the configurations
considered here.

\begin{figure}[!h]
    \centering
    \includegraphics[width=1.\linewidth]{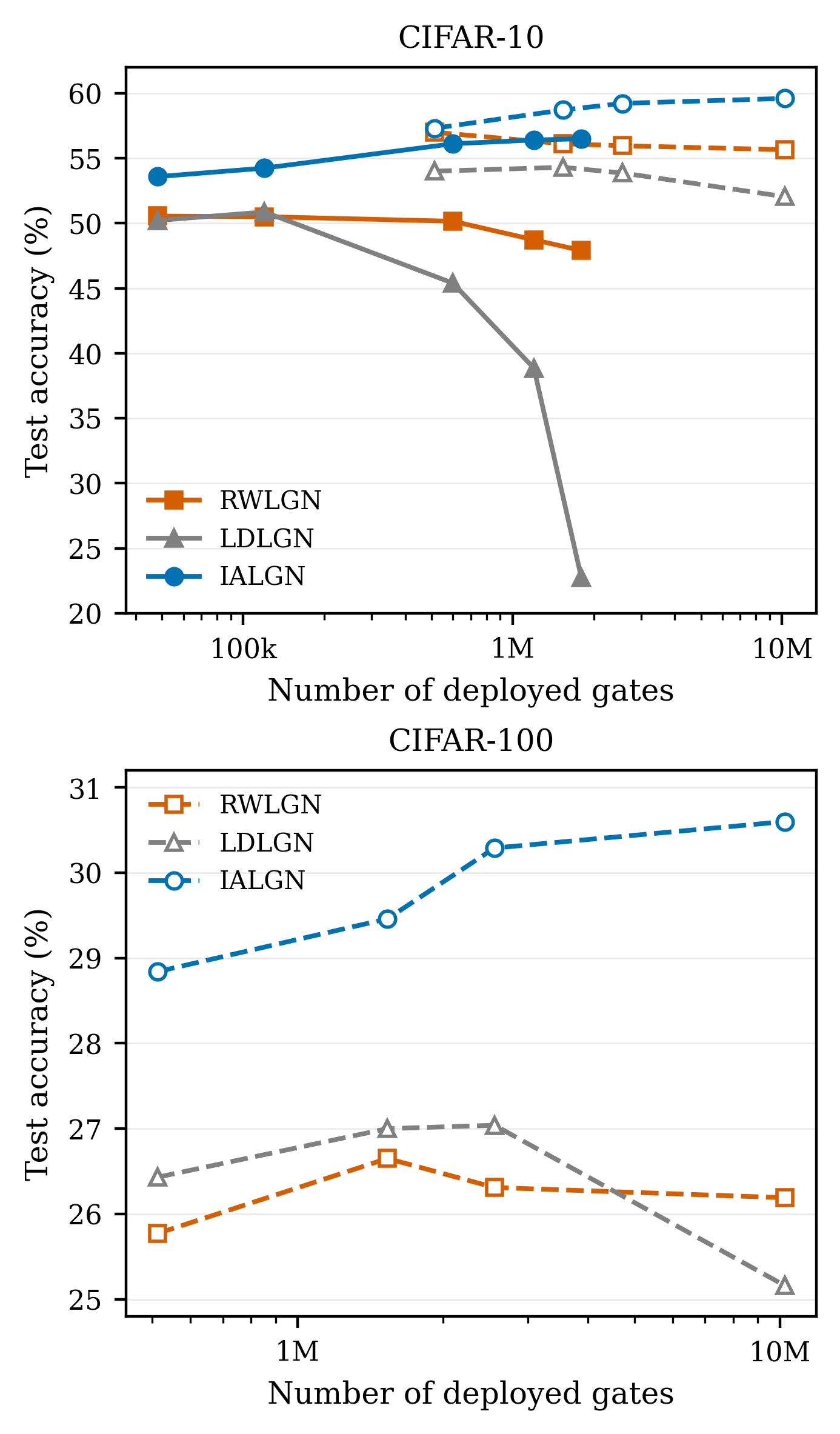}
    \caption{Test accuracy versus the number of deployed logic gates on CIFAR-10 and CIFAR-100. Each line connects a fixed-width depth sweep reported in Table~1 of the main paper, with deployed gate count computed as $W D$. In the CIFAR-10 panel, solid lines with filled markers denote $W=12k$, whereas dashed lines with hollow markers denote $W=128k$. CIFAR-100 uses $W=128k$. Across the evaluated configurations, IALGN provides a more favorable accuracy--gate-count trend, while RWLGN and LDLGN generally saturate or degrade as circuit size increases.}
    \label{fig:gate_count_accuracy}
\label{fig:gate_count_accuracy}
\end{figure}

\subsection{Input Binarization and Anchor Candidate Size.}

\begin{figure}[h]
    \centering
    \includegraphics[width=1.\linewidth]{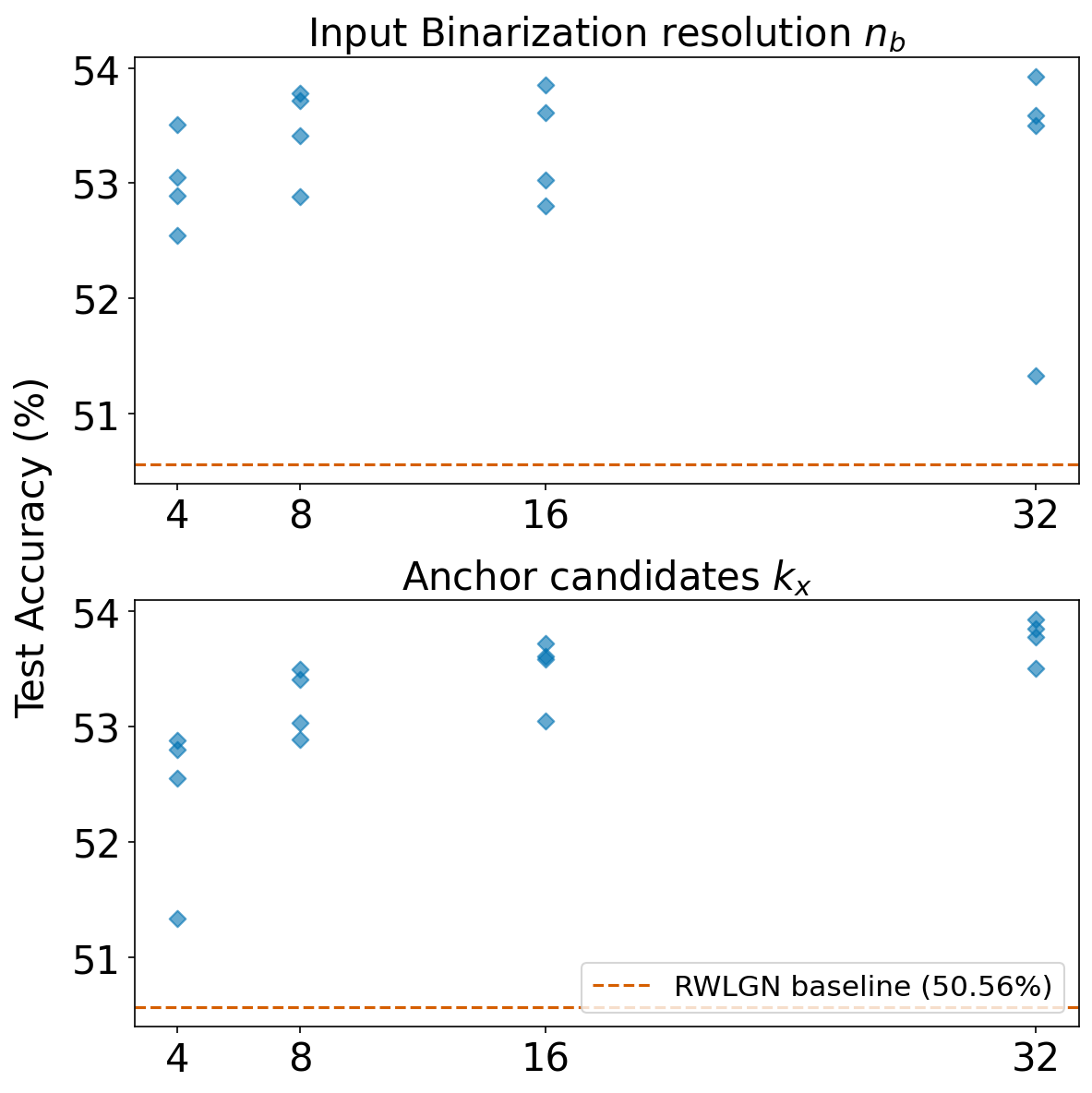}
    \caption{Effect of input encoding and anchor candidate size on CIFAR-10 at $W=12$k and $D=4$. Each marker represents one $(n_b,k_x)$ configuration. All variants use the same network width, depth, and deployed gate count. Increasing $k_x$ adds anchor-selection flexibility during training but does not increase the number of deployed gates at inference.}
    \label{fig:bk_ablation}
\end{figure}

We evaluate $n_b\in\{4,8,16,32\}$ and $k_x\in\{4,8,16,32\}$, with $W=12k$, $D=4$, and $k_h=1$. Each configuration is evaluated over three independent seeds. These hyperparameters affect how input information is represented and how flexibly each gate can select an input anchor, but they do not change the core input-anchored topology. Increasing $k_x$ introduces a diverse differentiable candidate set for the anchor pin, allowing each gate to select a more useful input feature without performing a global wiring search or increasing the deployed gate count.

Figure~\ref{fig:bk_ablation} summarizes the effects of $n_b$ and $k_x$ for shallow IALGN models. Increasing $k_x$ can improve performance when the candidate set contains informative input features, but the gains are not uniformly monotonic across all binarization resolutions. Input encoding exhibits a related trade-off: increasing $n_b$ provides a richer binary representation, but it also enlarges the input space from which anchors must be selected.

These results clarify the role of random-$k_x$ anchor relaxation. Input anchoring provides the architectural change associated with depth scaling, whereas random-$k_x$ relaxation improves anchor-selection flexibility within the anchored topology without changing the deployed gate count.

\section{Effective-Depth Analysis}
\label{sec:supp_effective_depth}

\begin{table}[h]
\centering
\small
\caption{Operation categories used in the effective-depth analysis.
Operation IDs follow the zero-based implementation ordering.}
\begin{tabular}{lll}
\toprule
Category & Operations & Depth update \\
\midrule
Constants
& $0,15$
& $0$ \\
Pass-through
& $3\;(A),\,5\;(B)$
& $d^A,\,d^B$ \\
Unary negation
& $12\;(\neg A),\,10\;(\neg B)$
& $d^A+1,\,d^B+1$ \\
Binary
& Remaining ten operations
& $1+\max(d^A,d^B)$ \\
\bottomrule
\end{tabular}
\end{table}

\paragraph{Operation-aware effective depth.}
We compute effective depth after discretizing Boolean-operation choices and, for IALGN, anchor selections by argmax. RWLGN hidden connections remain fixed throughout training. Raw input bits have depth zero. Let $d^A_{\ell,i}$ and $d^B_{\ell,i}$ denote the effective depths of the two inputs to gate $i$ at layer $\ell$. For RWLGN, both values are inherited from the corresponding randomly selected hidden predecessors. For IALGN, $A$ denotes the hidden spine and $B$ denotes the direct input anchor; hence $d^B_{\ell,i}=0$ after the first layer. Both inputs to the first layer have depth zero.

For a discretized operation $g_{\ell,i}$, effective depth is defined recursively as
\[
d_{\ell,i} =
\begin{cases}
0,
& g_{\ell,i}\in\{\mathrm{FALSE},\mathrm{TRUE}\},\\
d^A_{\ell,i},
& g_{\ell,i}=A,\\
d^B_{\ell,i},
& g_{\ell,i}=B,\\
d^A_{\ell,i}+1,
& g_{\ell,i}=\neg A,\\
d^B_{\ell,i}+1,
& g_{\ell,i}=\neg B,\\
1+\max\{d^A_{\ell,i},d^B_{\ell,i}\},
& \text{otherwise}.
\end{cases}
\]
Thus, constants terminate preceding computation, exact pass-through operations preserve the depth of the relevant input, unary negations add one operation to their relevant input, and genuinely binary operations extend the longer incoming path by one.

\paragraph{Aggregation across seeds.}
For each architecture and dataset, we analyze five independently trained 100-layer checkpoints. Within each seed, the mean, 10th percentile, 90th percentile, and maximum effective depth are computed over gates at each layer. Figure~5 of the main paper reports the averages of these seed-level statistics. In particular, the dashed curves average the per-seed maxima rather than taking a single global maximum after pooling gates across seeds. Operation frequencies are likewise computed separately for each seed and then averaged. The first layer is excluded from this census because both pins originate from raw inputs and do not yet have spine--anchor semantics.

\paragraph{Interpretation.}
This metric measures the maximum number of non-trivial Boolean operations along a surviving input-to-gate path. It is operation-aware but does not measure output-specificity or credit quality. In particular, an RWLGN may obtain a large effective depth by repeatedly taking the deeper parent within a branching and broadly shared ancestry. Effective depth must therefore be interpreted jointly with the path-sharing and credit-assignment diagnostics in the main paper.

\paragraph{Operation-frequency analysis.}
For operation-frequency analysis, we exclude the first layer because both of its pins are raw inputs and therefore do not yet have spine and anchor semantics. The remaining operations are grouped into constants, pass-through operations, unary negations, and the ten Boolean functions that genuinely depend on both inputs. In IALGN, $A$ preserves the spine, whereas $B$ and $\neg B$ discard the accumulated spine state and use only the direct input anchor.

\end{document}